\pdfoutput=1

\documentclass[a4paper,fleqn]{cas-dc}

\usepackage{lipsum}
\usepackage{lineno}
\usepackage{multirow}
\usepackage{graphicx}
\usepackage{grffile}
\usepackage{epstopdf}

\usepackage{upgreek}
\usepackage{bm}

\usepackage{ragged2e}
\usepackage{graphicx}
\renewcommand{\raggedright}{\leftskip=0pt \rightskip=0pt plus 0cm}

\makeatletter
\renewcommand\subsection{\@startsection{subsection}{2}{\z@}%
    {10pt \@plus 3\p@ \@minus 2\p@}%
    {.1\p@}%
    {%\let\@hangfrom\relax
     \ssectionfont\raggedright }}
\usepackage{amsmath,amsthm}
\usepackage{setspace}

\usepackage[numbers,,sort&compress]{natbib}
\usepackage{amsfonts}
\usepackage{amsmath}
\xspaceaddexceptions{]}
\def\tsc#1{\csdef{#1}{\textsc{\lowercase{#1}}\xspace}}
\tsc{WGM}
\tsc{QE}
\tsc{EP}
\tsc{PMS}
\tsc{BEC}
\tsc{DE}

\usepackage{subcaption}

\usepackage{natbib} % cas-cd require that this or similar bib package is loaded
\usepackage{lipsum} % 

\begin{document}
\let\WriteBookmarks\relax
\def\floatpagepagefraction{1}
\def\textpagefraction{.001}
\shorttitle{Training Neural Networks for Solving 1-D Optimal Piecewise Linear Approximation}
\shortauthors{Dong et al.}
%\begin{frontmatter}

\title [mode = title]{Training Neural Networks for Solving 1-D Optimal Piecewise Linear Approximation}                      
%%\tnotemark[1,2]

\author[1]{Hangcheng Dong}[orcid=0000-0002-4880-6762]
\credit{Harbin Institute of Technology}

\address[1]{School of Instrumentation Science and Engineering, Harbin Institute of Techonoloy, Harbin, 150001, China }

\author[1]{Jingxiao Liao}
\author[1]{Yan Wang}
\author[2]{Yixin Chen}
\author[1]{Bingguo Liu}
\author[1]{Dong Ye}
\author[1]{Guodong Liu \corref{cor1}}

\cormark[1]
\ead{lgd@hit.edu.cn}
%\cormark[1]
%\cortext[cor1]{Corresponding author}

\address[2]{BIOMIND}

\begin{abstract}[S U M M A R Y]
Recently, the interpretability of deep learning has attracted a lot of attention. A plethora of methods have attempted to explain neural networks by feature visualization, saliency maps, model distillation, and so on. However, it is hard for these methods to reveal the intrinsic properties of neural networks. In this work, we studied the 1-D optimal piecewise linear approximation (PWLA) problem, and associated it with a designed neural network, named lattice neural network (LNN). We asked four essential questions as following: (1) What are the characters of the optimal solution of the PWLA problem? (2) Can an LNN converge to the global optimum? (3) Can an LNN converge to the local optimum? (4) Can an LNN solve the PWLA problem? Our main contributions are that we propose the theorems to characterize the optimal solution of the PWLA problem and present the LNN method for solving it. We evaluated the proposed LNNs on approximation tasks, forged an empirical method to improve the performance of LNNs. The experiments verified that our LNN method is competitive with the start-of-the-art method. 

\end{abstract}
\begin{keywords}
Deep learning \sep Neural networks   \sep Interpretability \sep Piecewise linear models \sep Optimal approximation 
\end{keywords} 

\maketitle

\section{Introduction}

In recent years, as the mainstream machine learning model, deep learning has made a profound impact on many fields, including computer vision~\cite{he2016deep,vit,swin-transformer}, nature language processing~\cite{gpt3,xlnet,attention}, reinforcement learning~\cite{alphazero}, and so on. To pursue higher performance, deep learning practitioners tend to design networks with more complex architectures and a larger number of parameters~\cite{alexnet,vggnet,googlenet,densenet}. However, neural networks are criticized as black-box models, which are incurred by the opaqueness. The lack of interpretability has been the primary bottleneck of deep learning, impeding the widespread translation to mission-critical domains~\cite{lipton_mythos_2018}. 

To verify the decision process of neural networks, researches on the interpretability of deep learning have become an emerging focus~\cite{zhang_survey_2021,survey,XAI}. There exist two methodologies in the current discussions upon interpretability, one is the post-hoc explanation and the other is to construct explainable models~\cite{Fan2021}. In our previous work~\cite{work1}, we have discussed that the essence of interpretation is the completeness, which induces us to dig into the post-hoc explanation of models by finding their piecewise linear approximations that are easy-to-understand yet with powerful approximation ability. 

Beyond the post-hoc explanation, can we understand the training behavior of neural networks? It has been noticed that the infinite width fully connected neural networks could converge to the Gaussian process~\cite{NNGP,neal1996:priors}. However, it is difficult to characterize the optimal solution of networks with finite width. Motivated by the connection between ReLU NNs and piecewise linear functions, i.e., with the continuous piecewise linear activation function, the composite of fully connected layers is still continuous piecewise linear mapping~\cite{spline,bengio,chu_exact_2018}, we study the problem so-called one-dimensional optimal piecewise linear approximation in least square sense and associate it with a designed neural network, which promotes our understanding of the behavior of neural networks. 

H.Stone~\cite{Opa1961} introduced the 1-D PWLA problem and showed that there is no closed-form solution. Golovchenko~\cite{CPWL_fixed2004} analyzed the continuous PWLA problem with fixed break points could be handled by least squares fit. R.Bellman et al.~\cite{1961bellman} proposed a sequence search method by transforming the n-segment PWLA problem into (n-1)-segment one. Cleghorn et al.~\cite{2012particle} reported a particle swarm optimization method for n-dimensional PWLA. Jekel et al.~\cite{2019pwlf} released a Python library named pwlf for fitting 1-D continuous piecewise linear functions, in which the differential evolution method had been used for the global optimization.

In this paper, we designed a one-hidden neural network called lattice neural network for seeking the optimal solution of the 1-D continuous PWLA (CPWLA) problem. Our main contributions are summarized as follow:

(1) We obtained the necessary and sufficient conditions of (C)PWLA problems.

(2) We proposed a lattice neural network and associated it with the CPWLA problem. Moreover, we explored the properties of the optimum of LNNs according to the theorem proposed above, which shed a light on the understanding of neural networks. 

(3) We designed an empirical method to improve the performance of LNNs.

\section{Problem Definition}

For a given one-dimensional nonlinear relationship $y=f(x)$, $\mu_0 \leq x \leq \mu_n$, it is desired to obtain an approximation of the piecewise linear form :

\begin{equation}
    \label{eq1}
    g(x)=\begin{cases}
    a_1+b_1x & \mu_0 \leq x < \mu_1 \\
    a_2+b_2x & \mu_1 \leq x < \mu_2 \\
    ...      & ...               \\
    a_n+b_nx & \mu_{n-1} \leq x \leq \mu_n
    
    \end{cases}
\end{equation}
which minimizes the least squares error (LSE) between $f(x)$ and $g(x)$ on the known interval $x \in [\mu_0,\mu_n]$, namely:

\begin{equation}
    \label{eq2}
    \begin{aligned}
    \text{LSE}=E_n(x) &=\int ^{\mu_n} _{\mu_0} (f(x)-g(x))^2 dx \\
    &= \sum ^n_{i=1} \int ^{\mu_i}_{\mu_{i-1}} (f(x)-g_i(x))^2 dx
    \end{aligned}
\end{equation}
where $a_i$, $b_i$ are respectively the intercepts and slopes of the linear function $g_i(x)=a_i+b_ix$ on the interval $x\in [\mu_{i-1},\mu_i)$ ($[\mu_{n-1},\mu_n]$), $i=1,2,...,n$. It is worth to noticed that the parameters $\mu_i$, for $i=1,2,...n-1$ ($\mu_0$, $\mu_n$ are fixed), are also the variables of the function $g(x)$.

For convenience, if there are n segments of $g(x)$, we call it n-order least squares piecewise linear approximation, abbreviated as n-order PWLA. Furthermore, if $g(x)$ is $C_0$ continuous, we abbreviate it as n-order CPWLA.

%%%%%
%%%%%
%%%%%

\section{Properties of (C)PWLA}

First, we define some auxiliary functions and notations. Then, we present two theorems to describe the properties of the solution of (C)PWLA and prove them respectively.\\

\textbf{Error function} We define the error function between $f(x)$ and $g(x)$ on the interval $<\alpha,\beta>$ (it dose not matter that the interval is open or close, as long as it does not degenerate to a point) as:

\begin{equation}
    \label{eq3}
    e(\alpha,\beta)=\int^{\beta}_{\alpha} (f(x)-g(x))^2dx
\end{equation}

It is obvious that $e(\alpha,\beta)$ is the monotone increasing (decreasing) differentiable function about $\beta$ ($\alpha$), and then the Eq.(\ref{eq2}) can be rewritten as $E_n(x)=\sum^{n} _{i=1} e(\mu_{i-1},\mu_i)$.\\

\textbf{Notations of the n-order (C)PWLA} Denote by $H_n$ all the piecewise linear functions with n segments on the interval $[\mu_0,\mu_n]$, ($n\in \mathbb{Z}^+$, $\mu_0,\mu_n \in \mathbb{R}$). Denote by $f(x)$ the target function, namely the approximated function of (C)PWLA on the interval $[\mu_0,\mu_n]$. And the approximate function are noted as $g(x)\in H_n$, while the linear function on each interval $I_i=<\mu_{i-1},\mu_i>, (i=1,2,...,n)$ are given as $g_i(x)$. Then, the optimal solution of n-order (C)PWLA can be written as:

\begin{equation}
    \label{eq4}
    g^{*}(n)=\mathop{\arg\min}_{g(x)\in H_n} \int ^{\mu_n} _{\mu_0} (f(x)-g(x))^2 dx 
\end{equation}
In consequence, the minimum value of error function $E_n(x)$ will be: $E_{n} ^{*}(x)=\mathop{inf}_{g(x)\in H_n} \int ^{\mu_n} _{\mu_0} (f(x)-g(x))^2 dx$.

\noindent\textbf{Property 1} If $n_1\leq n_2$, then $E_{n_1}^{*}(x)\leq E_{n_2}^{*}(x)$.\\

Now, we represent the main results as follow.

\newtheorem{theorem}{Theorem}
\begin{theorem}\label{th1}
The function $g^{*}(n)$ is the optimal solution of the n-order PWLA, if and only if the function $g_i(x)=a_i+b_ix$ is the optimal solution of the 1-order PWLA on each linear interval $I_i=[\mu_{i-1},\mu_i]$ ($i=1,2,...,n$), and at each breakpoint $x=\mu_i$ ($i=1,2...,n-1$), $g^{*}(n)$ is either continuous, namely $g_{i-1}(\mu_i)=g_{i}(\mu_i)$, or satisfies $g_{i-1}(\mu_i)+g_{i}(\mu_i)=2f(\mu_i)$.
\end{theorem}

\newproof{pot}{Proof of Theorem \ref{th1}}
\begin{pot}
For the n-order PWLA, once all the breakpoints are fixed, let the $g_i(x)=a_i+b_ix$ take the least square solution, then we solve the problem. Thus, the problem is transformed into the search for the best breakpoints.For the breakpoint $\mu_p,(p=1,2,...,n-1)$, the optimal solution $g^{*}(n)$ should meet the following equation:
\begin{equation}
    \label{eq5}
     \frac{ \partial \sum ^n _{i=1} e(\mu_{i-1},\mu_i)}{\partial \mu_p}=0
\end{equation} 
By plugging $\frac{\partial e(\mu_{p-1},\mu_p)}{\partial \mu_p}=(f(\mu_p)-a_p-b_p\mu_p)^2$ and $\frac{\partial e(\mu_{p},\mu_{p+1})}{\partial \mu_p}=-(f(\mu_p)-a_{p+1}-b_{p+1}\mu_p)^2$ into above equation, we have:

\begin{equation}
    \label{eq6}
     (f(\mu_p)-a_p-b_p\mu_p)^2=(f(\mu_p)-a_{p+1}-b_{p+1}\mu_p)^2
\end{equation} 
which means that $g_{p-1}(\mu_p)=g_{p}(\mu_p)$ or $g_{p-1}(\mu_p)+g_{p}(\mu_p)=2f(\mu_p)$ for $p=1,2,...,n-1$.

\end{pot}

\begin{theorem}\label{th2}The function $g^{*}(n)$ is the optimal solution of the n-order CPWLA, if and only if the function $g_i(x)=a_i+b_ix$ is the optimal solution of the 1-order CPWLA on each linear interval $I_i=[\mu_{i-1},\mu_i]$ ($i=1,2,...,n$).
\end{theorem}

\newproof{pot2}{Proof of Theorem \ref{th2}}
\begin{pot2}
According to the continuity, we rewrite the Eq.(\ref{eq1}) as:

\begin{equation}
    \label{eq7}
    g(x)=\begin{cases}
    a_1+b_1x & \mu_0 \leq x < \mu_1 \\
    a_1+(b_1-b_2)\mu_1+b_2x & \mu_1 \leq x < \mu_2 \\
    ...      & ...               \\
    a_1+\sum^{n-1} _{i=1} (b_i-b_{i+1})\mu_i +b_nx & \mu_{n-1} \leq x \leq \mu_n
    
    \end{cases}
\end{equation}

Similar to Eq.(\ref{eq5}), the optimal solution $g^{*}(n)$ should meets following $2n$ equations:

\begin{equation}
    \label{eq8}
    \begin{cases}
     \sum ^n _{i=1} \frac{\partial e(\mu_{i-1},\mu_i)}{\partial b_p}=0 & p=1,2,...,n \\
    \sum ^n _{i=1} \frac{\partial e(\mu_{i-1},\mu_i)}{\partial \mu_q}=0 & q=1,2,...,n-1 \\
     \sum ^n _{i=1} \frac{\partial e(\mu_{i-1},\mu_i)}{\partial a_1}=0 &  \\
    \end{cases}
\end{equation}

Consider the following relationship:

\begin{equation}
    \label{eq9}
    \frac{\partial e_i}{\partial \mu_q}=\begin{cases}
     0 & q>i \\
    (f(\mu_q)-g_q(\mu_q)^2 & q=i \\
    -2(b_q-b_{q+1})A_i-(f(\mu_{q})-g_{q+1}(\mu_q)^2 & q=i-1 \\
    -2(b_q-b_{q+1})A_i & q<i-1 
    \end{cases}
\end{equation}
where $A_i=\int ^{\mu_i} _{\mu_{i-1}} (f(x)-g_i(x)) dx$ and $e_i=e(\mu_{i-1},\mu_i)$, then we can rewrite the $\sum ^n _{i=1} \frac{\partial e(\mu_{i-1},\mu_i)}{\partial \mu_q}=0$ $(q=1,2,...,n-1)$ as following:

\begin{equation}
    \label{eq10}
    \begin{cases}
     (b_1-b_2)\sum ^n _{k=2} A_k=0  \\
    (b_2-b_3)\sum ^n _{k=3} A_k=0 \\
    ... \\
    (b_{n-1}-b_n) A_n=0
    \end{cases}
\end{equation}

By simplifying the equation $\sum ^n _{i=1} \frac{\partial e(\mu_{i-1},\mu_i)}{\partial a_1}=0$, we have:

\begin{equation}
    \label{eq11}
    \sum ^n _{k=1} A_k=0
\end{equation}

By considering the Eq.(\ref{eq10}) and Eq.(\ref{eq11}), we have:
\begin{equation}
    \label{eq12}
    A_i=0  \quad  i=1,2,...,n 
\end{equation}

By the same way, we have:

\begin{equation}
    \label{eq13}
    \frac{\partial e_i}{\partial b_p}=\begin{cases}
     0 & p>i \\
    -2B_1 & p=i=1 \\
    -2(B_i-\mu_{p-1}A_i) & p=i \neq 1\\
    -2(\mu_p-\mu_{p-1})A_i & p<i 
    \end{cases}
\end{equation}
where $B_i=\int ^{\mu_i} _{\mu_{i-1}} (f(x)-g_i(x))x dx$. By combining Eq.(\ref{eq8}), Eq.(\ref{eq12}) and Eq.(\ref{eq13}), we have:
\begin{equation}
    \label{eq12}
    B_i=0  \quad  i=1,2,...,n 
\end{equation}
which means that $g_i(x)$ is the least squares solution of $f(x)$ on each interval $I_i=[\mu_{i-1},\mu_i]$ ($i=1,2,...,n$).

\end{pot2}

%%%%%%%%%%%%%%%%%%%%%
%%%%%%%%%%%%%%%%%%%%%
%%%%%%%%%%%%%%%%%%%%%
\section{Lattice Neural Network for CPWLA}

%\subsection{framework}
Considering a one-hidden-layer neural network with the one-dimensional input $x\in [\mu_0,\mu_n]$ and output $g(x)$, we design an activation function $\sigma(x)$ based on the lattice theory, whose formulation is: 

\begin{equation}
    \label{eq15}
    \sigma(x)=max(w_1x+b_1,w_2x+b_2)
\end{equation}

\begin{figure}
    \centering
    \includegraphics[scale=.45]{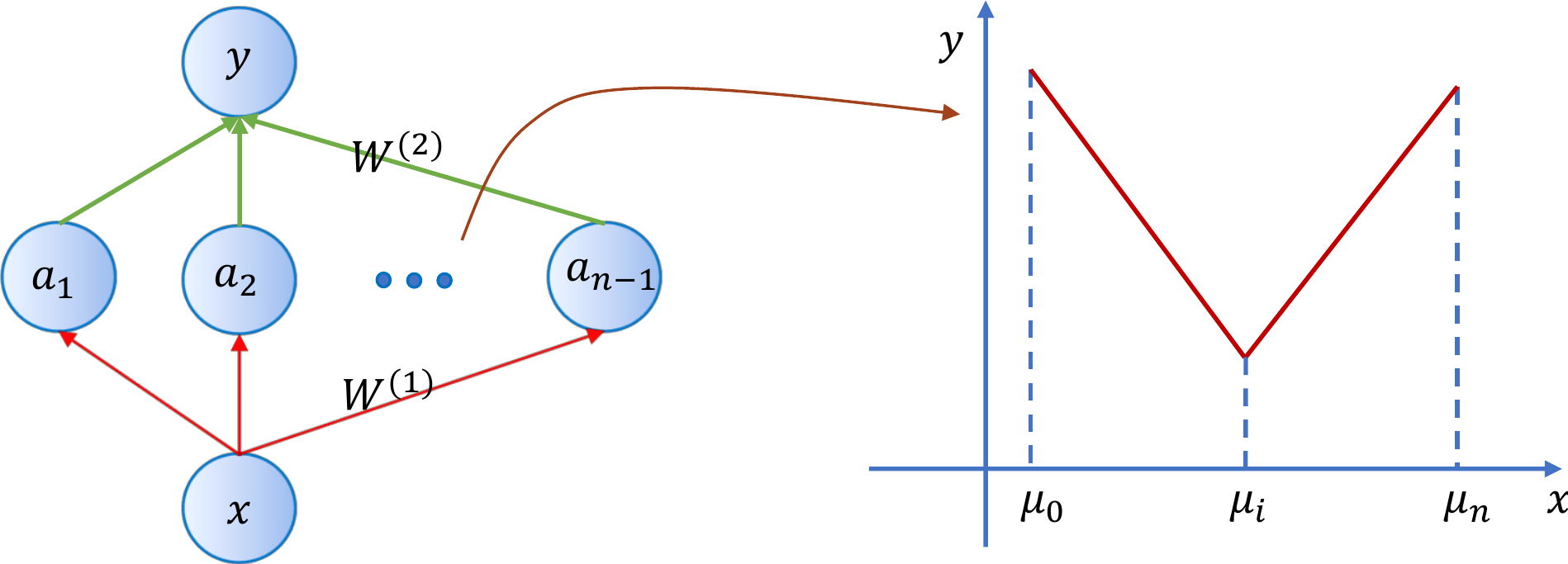}
    \caption{Architecture of the lattice neural network. Each hidden neuron corresponds to a 2-segment piecewise linear function.}
    \label{fig:1}
\end{figure}

\begin{figure}
    \centering
    \includegraphics[scale=.25]{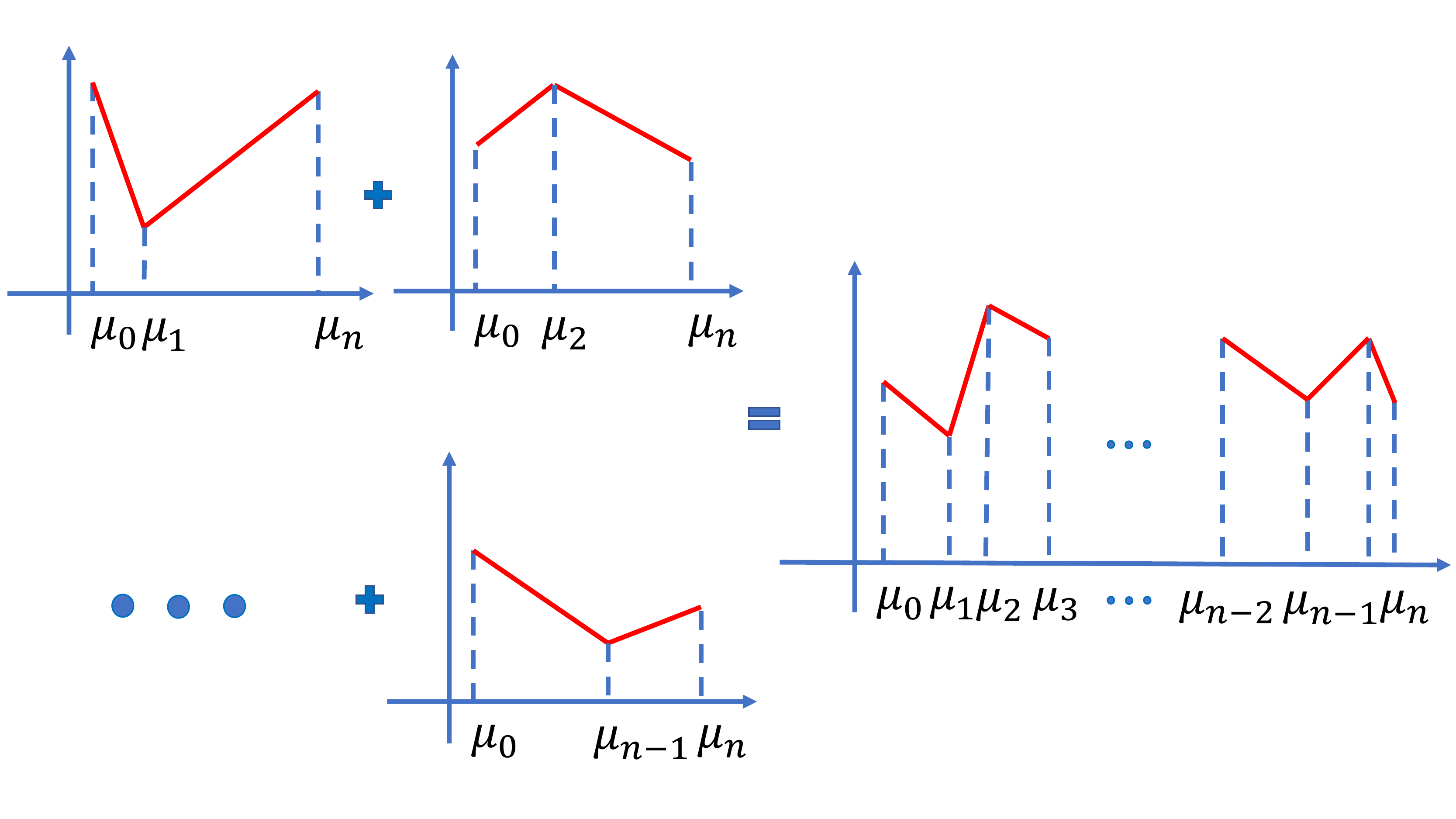}
    \caption{The sum of the output of the hidden layer with n-1 neurons in the LNN, which belongs to $H_n$}
    \label{fig:2}
\end{figure}

By taking the above activations, there is geometry interpretation for the lattice neural network (LNN, see Fig.\protect\ref{fig:1})), namely a single neuron can express any piecewise linear function with 2 segments. Therefore, as illustrated in the Fig.\protect\ref{fig:2}), each neuron corresponds to a breakpoint in the input space, and in consequence, the optimal solution of the LNN with n-1 hidden neurons should be the same as the n-order CPWLA when the loss function is taken MSE, namely:

\begin{equation}
    \label{eq16}
    g(x \vert \boldsymbol{\theta})=\arg \mathop{\min}_{\boldsymbol{\theta}} \frac{1}{m} \sum^{m}_{i=1} (f(x_i)-g(x_i \vert \boldsymbol{\theta} ))^2
\end{equation}
where $f(x)$ is the same target function of the n-order CPWLA, $\boldsymbol{\theta}$ represents all the parameters of the LNN and $m$ is the amount of the samples.

However, it is difficult to train the origin LNN to converge to the optimal solution of the n-order CPWLA, because the breakpoints tend to shift outside the interval $[\mu_0,\mu_n]$. 

To deal with the breakpoint shift problem, we limit the parameters and rewrite Eq.(\ref{eq15}) as follow:
{\small
\begin{equation}
    \label{eq17}
    \sigma(x)=max(w_1(x-(\frac{\mu_n-\mu_0}{1+exp(-\delta)}+\mu_0)),w_2(x-(\frac{\mu_n-\mu_0}{1+exp(-\delta)}+\mu_0)))
\end{equation}
}
where $\delta \in \mathbb{R}$. Thus, the breakpoints are limited in $[\mu_0,\mu_n]$. By the back-propagation algorithm~\cite{bp1986}, the LNN can be used for solve the optimal solution of the n-order CPWLA.

%%%%
%%%%
%%%%
\section{Experiments}

We evaluate the performance of the proposed LNN, and compare it with the pwlf algorithm~\cite{2019pwlf} in terms of accuracy and speed on the function approximation task. In particular, we address the following questions: (1) What does the optimal solutions of (C)PWLA look like? (2) Can LNNs approach the optimal solutions of CPWLA? (3) What are the advantages of LNNs for solving the CPWLA? (4) Does LNNs approach the best solution with the fixed? breakpoints? 

As shown in Table 1, we design three target functions to validate questions (1) and (2). To compare with the pwlf method better, we further designed another three test functions as listed in Table 2 which reference the experiments in \cite{2012particle}. To apply our LNN algorithm and pwlf method, we sample a grid of equally spaced points in the given intervals, whose quantity is taken by $m=2000$. 

\begin{table}
  \caption{Validation target functions.}
  \begin{tabular*}{\tblwidth}{@{} LLLL@{} }
   \toprule
    Functions & Intervals\\
   \midrule
    $f(x)=x^2$ & [-1,1]\\
    $f(x)=x^3$ & [-1,1]\\
    $f(x)=5xsin(5x) + cos(5x)sin(10x)+exp(-x)$ & [-1,1]\\
   \bottomrule
  \end{tabular*}
\end{table}

\begin{table}
  \caption{Test target functions.}
  \begin{tabular*}{\tblwidth}{@{} LLLL@{} }
   \toprule
    Functions & Intervals\\
   \midrule
    $f(x)=sin(\pi x)/{\pi x}$ & [-4,4]\\
    $f(x)=sin(x)+xsin(x)cos(x)$ & [-10,10]\\
    $f(x)=20-5exp(-0.3x)-3exp(cos(\pi x))$ & [-6,6]\\
   \bottomrule
  \end{tabular*}
\end{table}
 
We use Pytorch to implement our algorithm and the Python code of pwlf is publicly available. All the experiments were running on a PC with a Core-i7-8700k CPU (3.70 GHz) and 16 GB of main memory.

\subsection{What does the optimal solutions of (C)PWLA look like?}

In this experiment, we verify the proposed Theorem \ref{th1} and \ref{th2} with the result depicted. 

\begin{figure}
    \centering
    \setkeys{Gin}{width=\linewidth,height=3cm} %set image parameters
\begin{subfigure}{4cm}
    \includegraphics{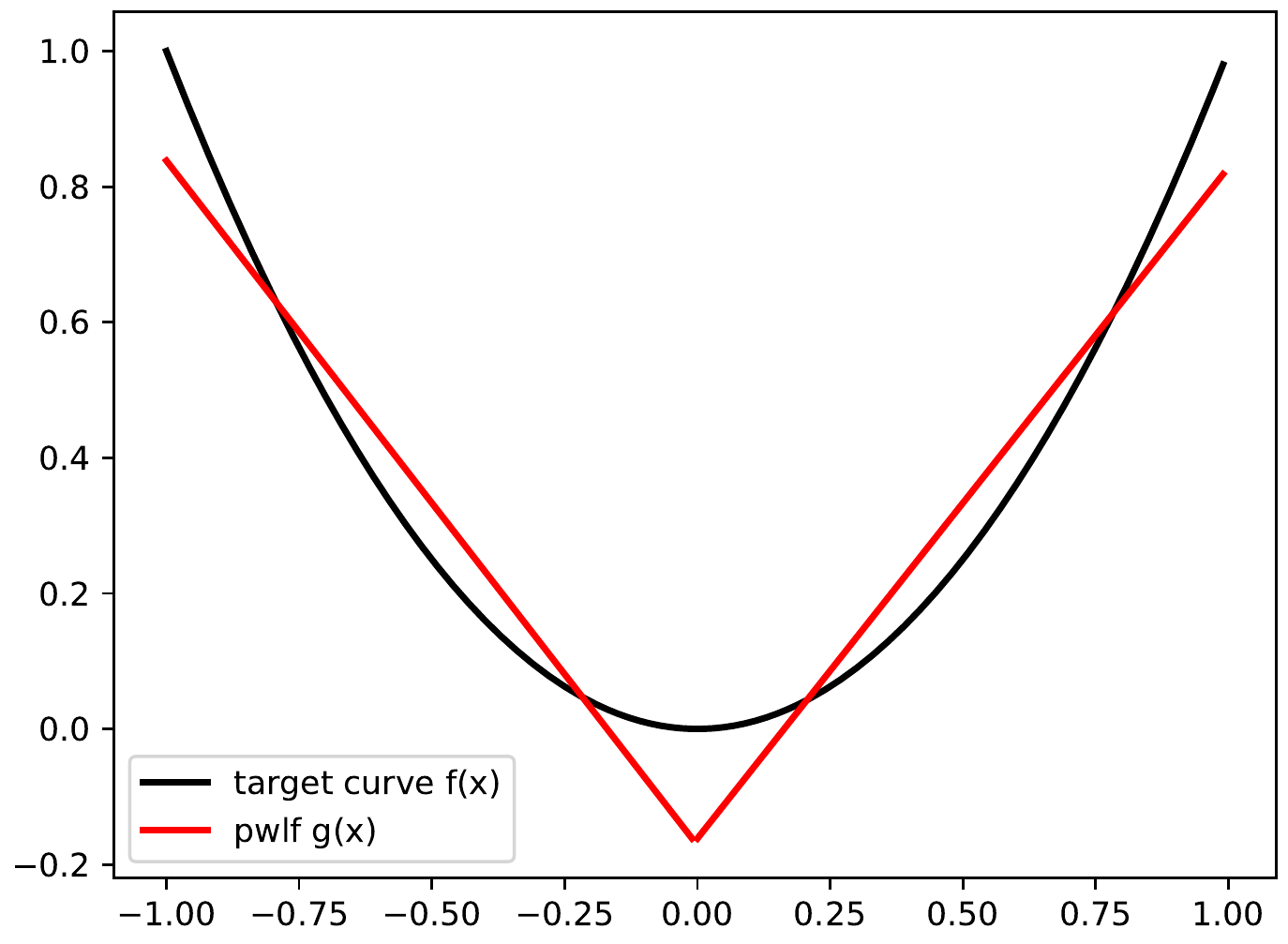}
    \caption{2-order PWLA, $y=x^2$}
    \label{fig3a}
\end{subfigure}
\hfil
\begin{subfigure}{4cm}
    \includegraphics{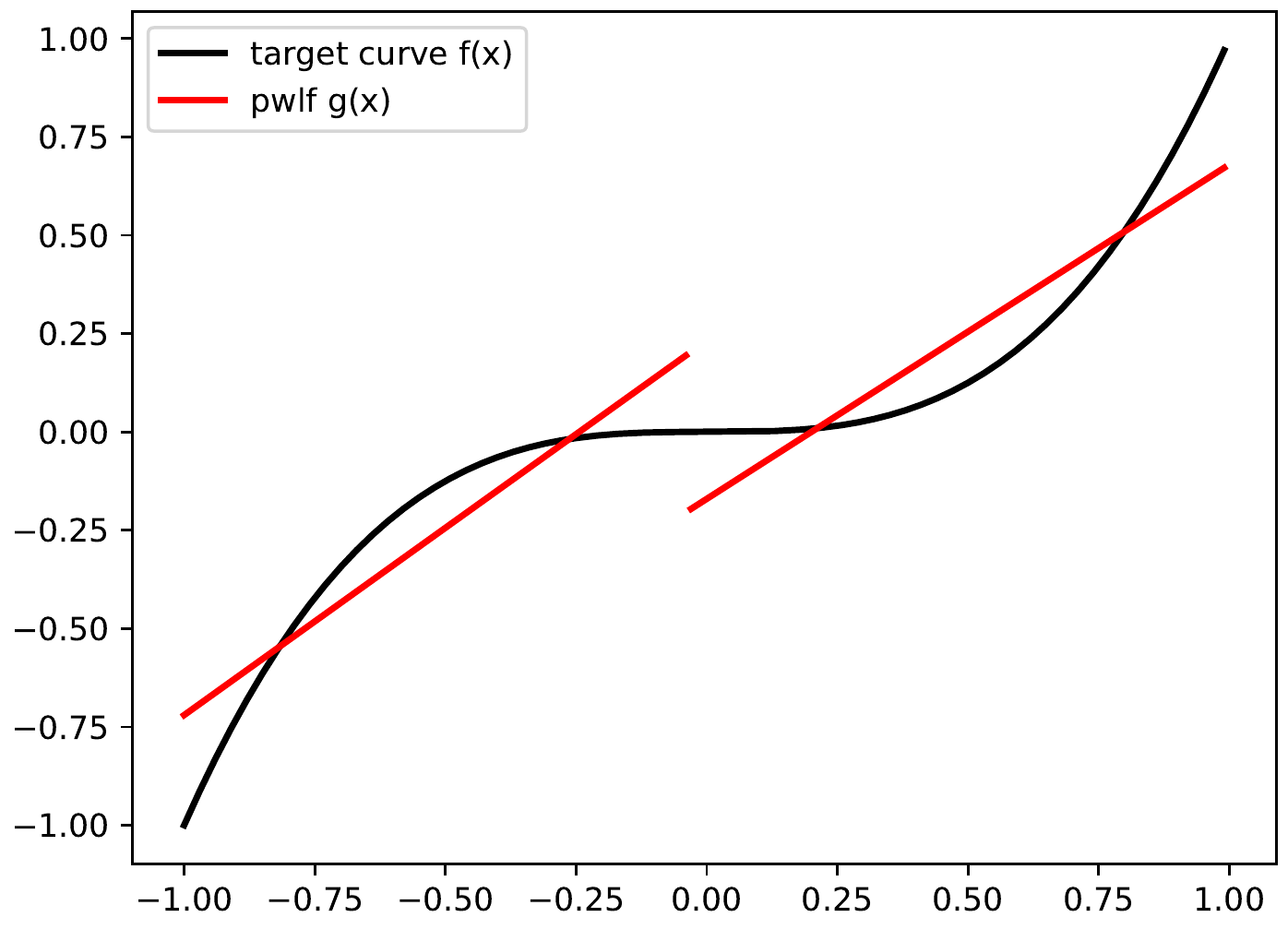}
    \caption{ 2-order PWLA, $y=x^3$}
    \label{fig3b}
\end{subfigure}

\medskip
\begin{subfigure}{4cm}
    \includegraphics{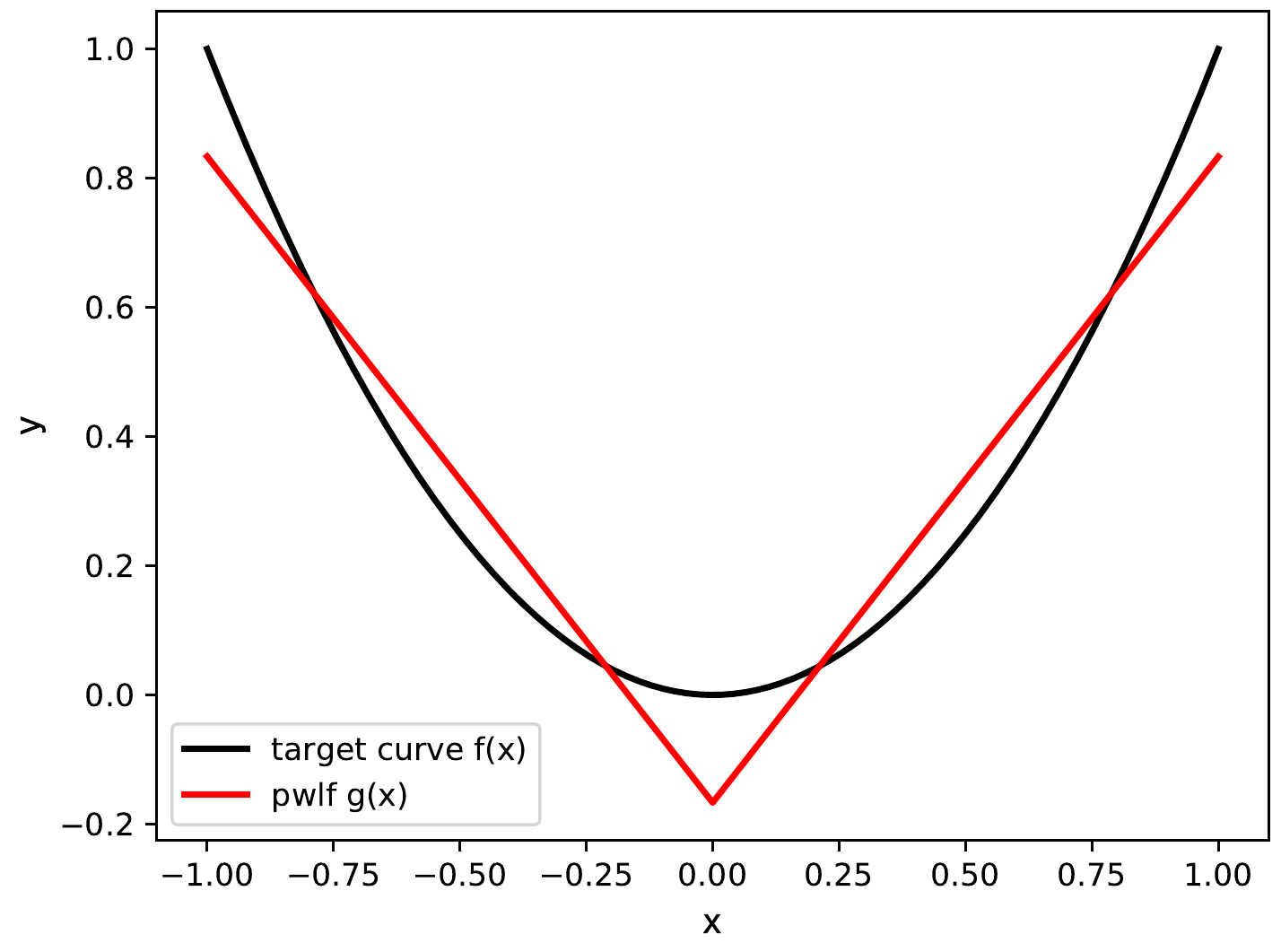}
\caption{2-order CPWLA, $y=x^2$}
\label{fig3c}
\end{subfigure}
\hfill
\begin{subfigure}{4cm}
    \includegraphics{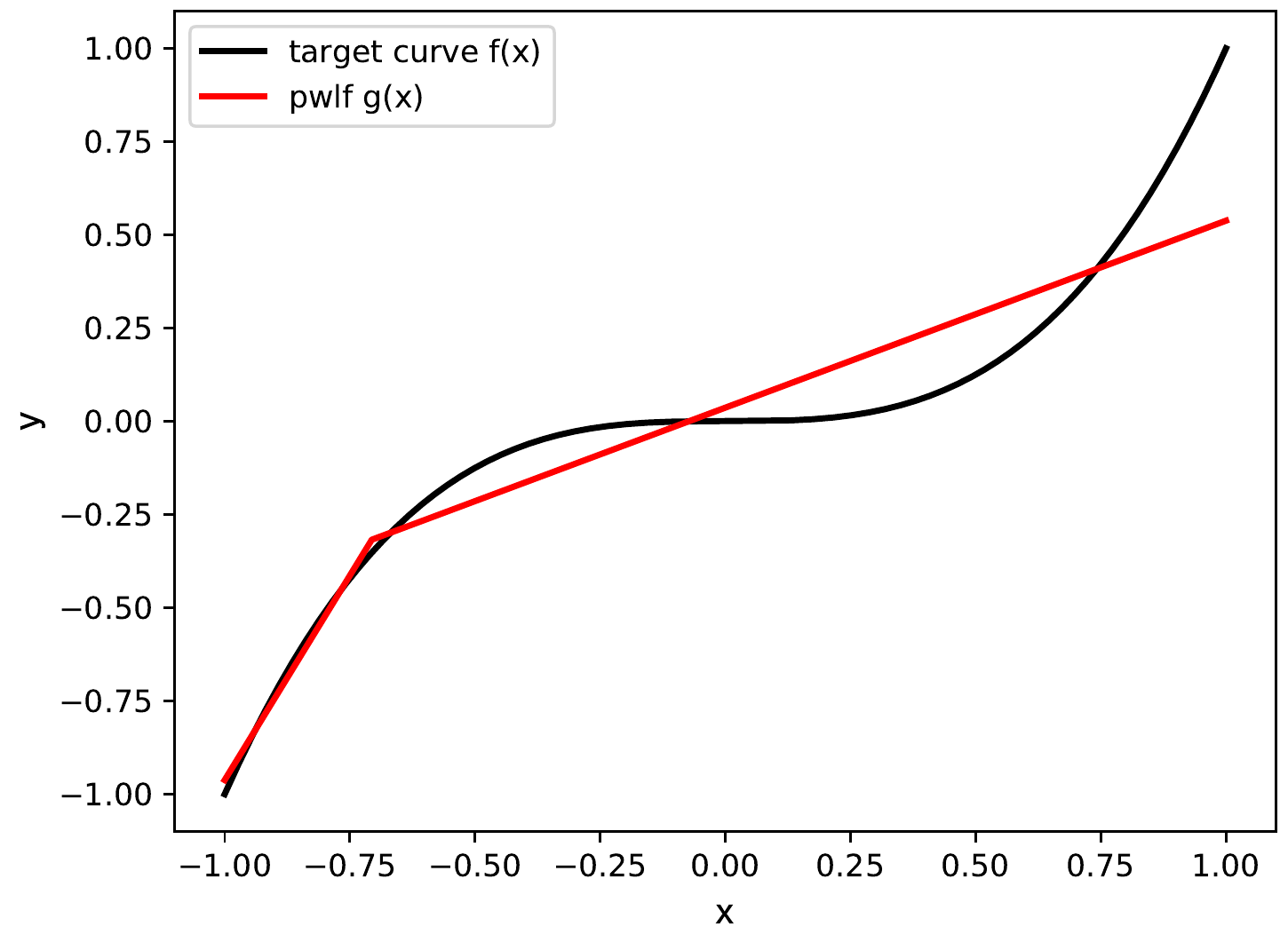}
\caption{2-order CPWLA, $y=x^3$}
\label{fig3d}
\end{subfigure}

\caption{The optimal solutions of 2-order (C)PWLA, (a)-(b) and (c)-(d) show the optimum of PWLA and CPWLA, respectively. The target function of (a) and (c) is $y=x^2$. The target function of (b) and (d) is $y=x^3$. }
\label{fig3}
    \end{figure}

We set the target functions as $f(x)=x^2$ and $f(x)=x^3$ on the interval $I=[-1,1]$, and contrast their optimal solutions of 2-order PWLA and CPWLA. Figures \ref{fig3a}-\ref{fig3b} show the results of PWLA, and figures \ref{fig3c}-\ref{fig3d} display those of CPWLA. As shown in Fig.\ref{fig3}, the optimal solutions of PWLA, which are approximately obtained by scanning the 2000 isometric points on interval I, exhibit different properties on curves with different concavities. In addition, it is obvious that there exist two optimal solutions of $y=x^3$, ($x\in [-1,1]$), and they are symmetrical about the zero point.

\begin{figure}
    \centering
    \setkeys{Gin}{width=8cm,height=6cm}
    \includegraphics{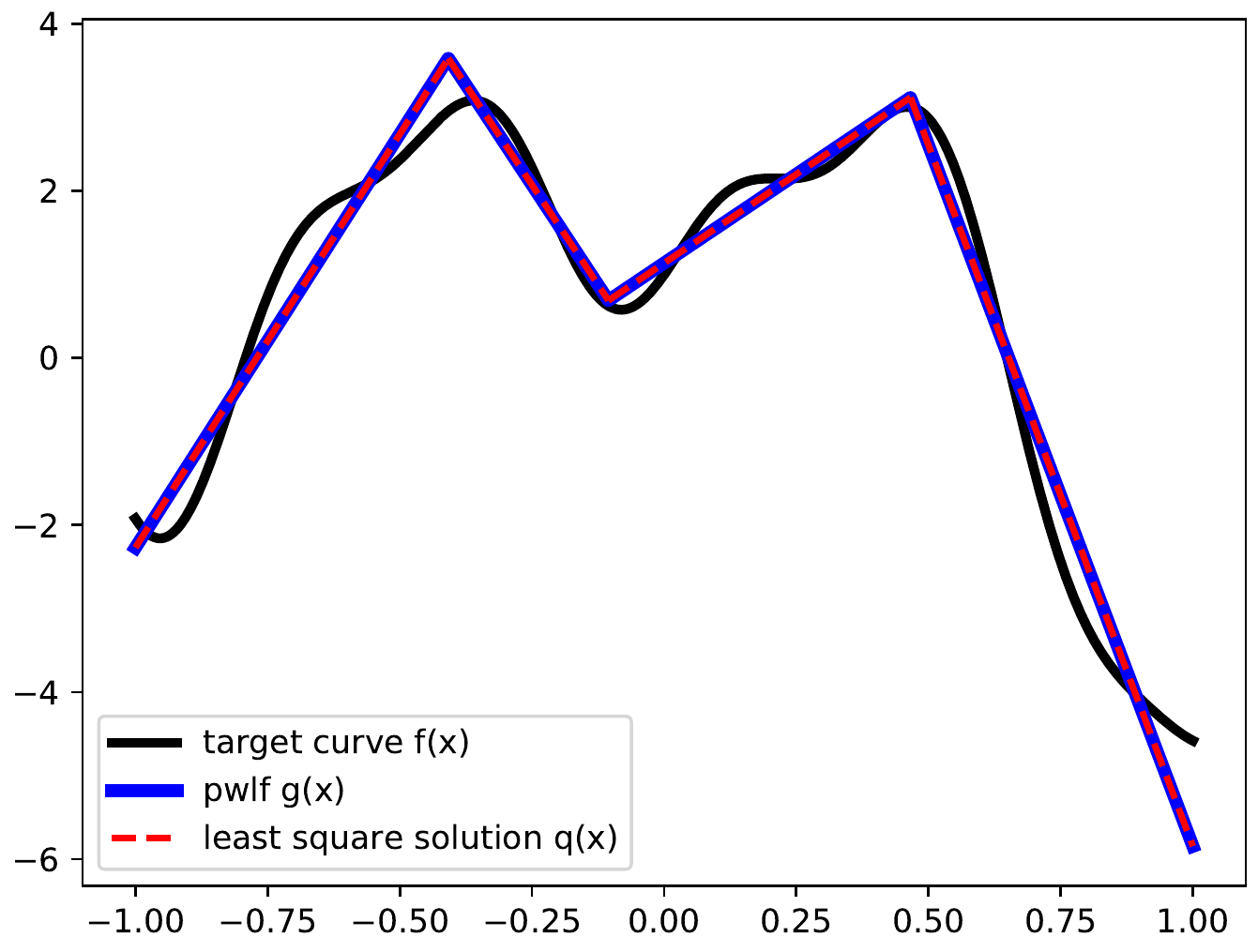}
    \caption{The optimal solutions of 4-order CPWLA. The solid blue line is the result solving by the pwlf, and the dashed red line is the least squares solution on each subinterval.}
    \label{fig:4}
\end{figure}

In Fig.\ref{fig:4}, we use the algorithm pwlf~\cite{2019pwlf} to compute the optimal solution $g(x)$ of the 4-order CPWLA with the target function set as $f(x)=5xsin(5x) + cos(5x)sin(10x)+exp(-x)$ on interval $[-1,1]$. To verify this, we calculate the least squares approximations of the target function on each of the four subintervals respectively. As it is shown, the dashed red lines (least square solutions) exactly match the solid blue line (the solution searched by the pwlf), which suggests that the pwlf successfully found the optimal solution.

%\lipsum[11]

\subsection{Can LNNs approach the optimal solutions of CPWLA?}

\begin{figure}
    \centering
    \setkeys{Gin}{width=\linewidth,height=6cm} %set image parameters
\begin{subfigure}{8cm}
    \includegraphics{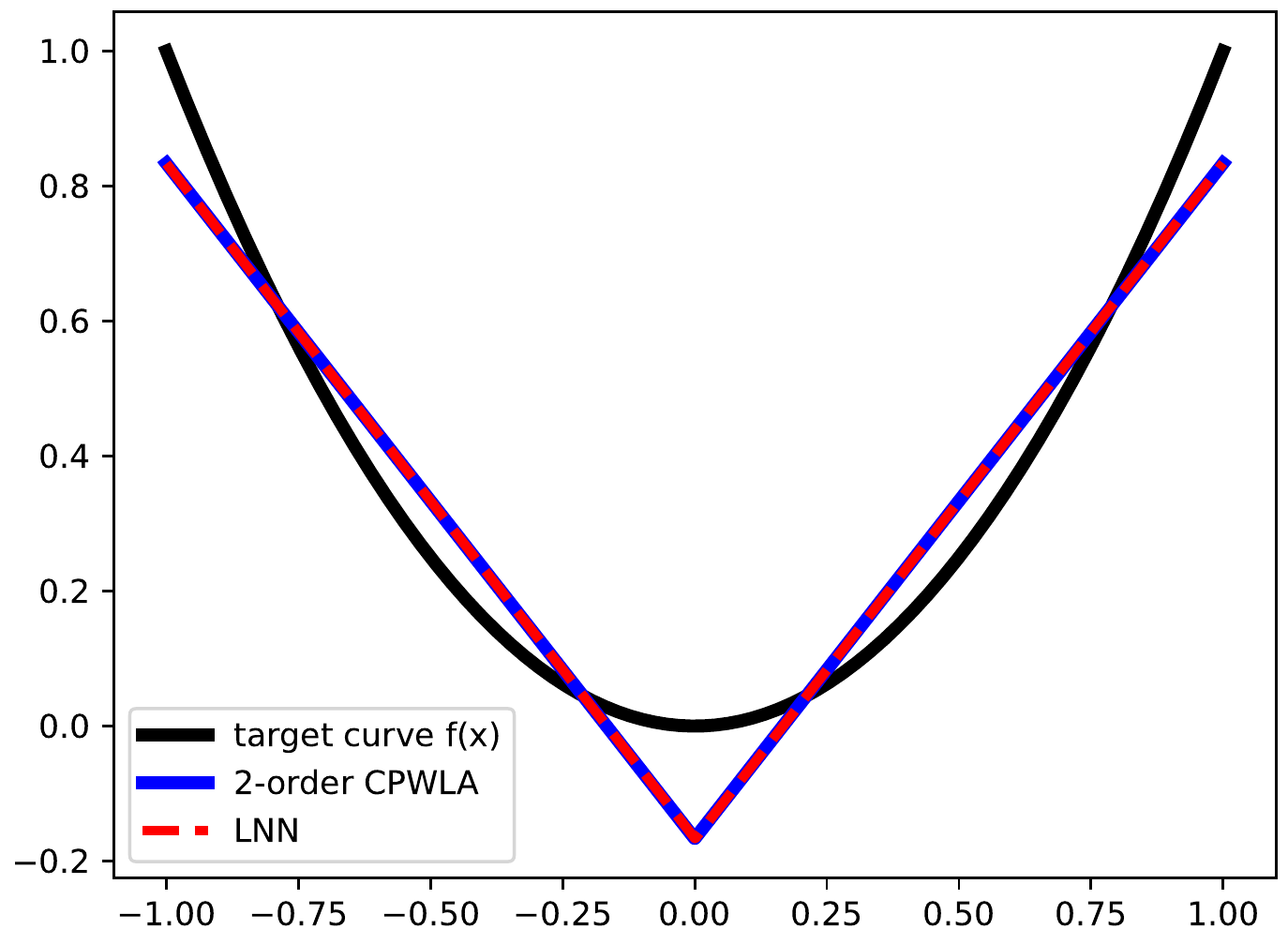}
    \caption{$y=x^2, x\in [-1,1]$}
    \label{fig5a}
\end{subfigure}
\hfil
\begin{subfigure}{8cm}
    \includegraphics{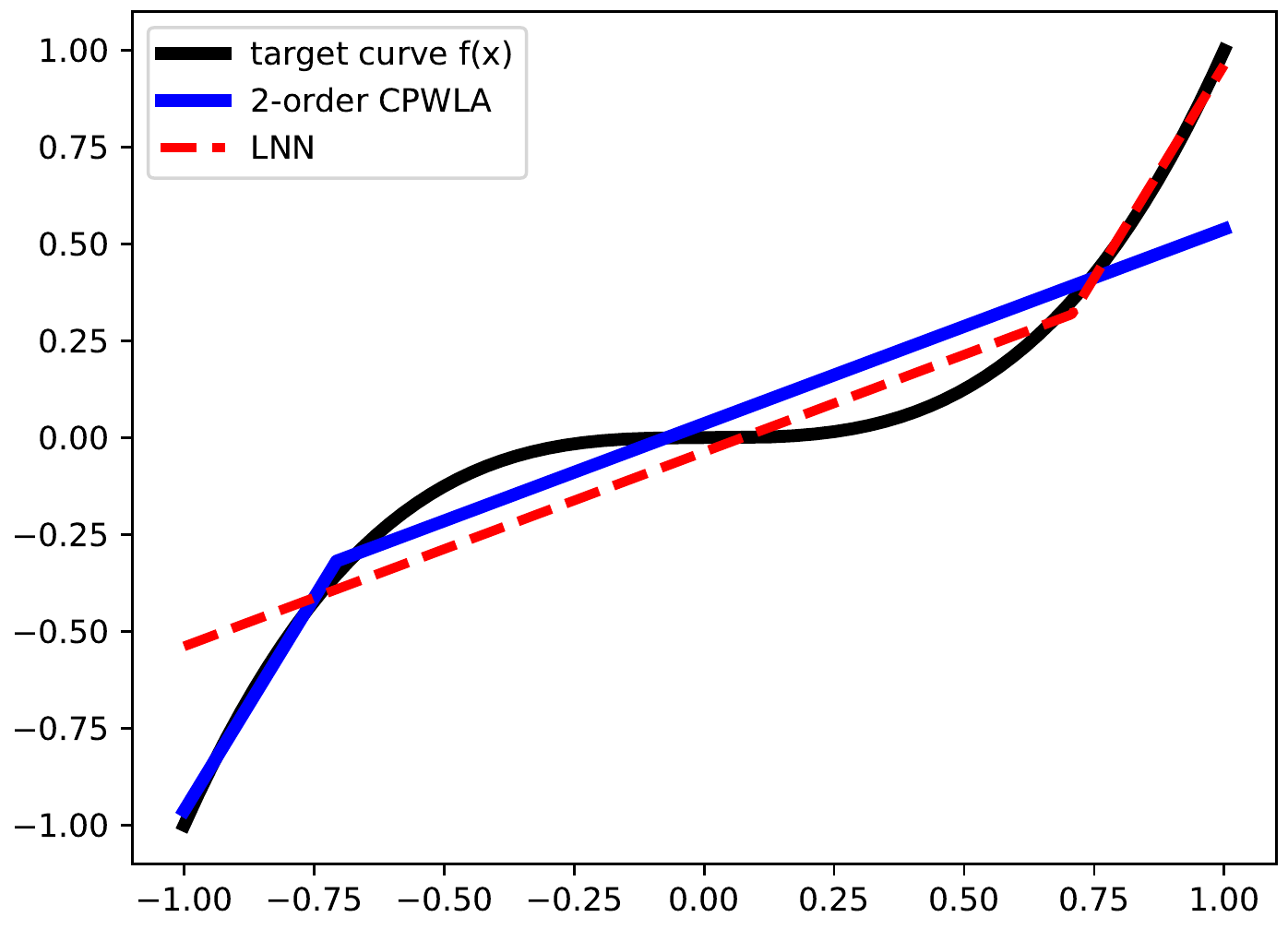}
    \caption{ $y=x^3, x\in [-1,1]$ }
    \label{fig5b}
\end{subfigure}

\medskip
\begin{subfigure}{8cm}
    \includegraphics{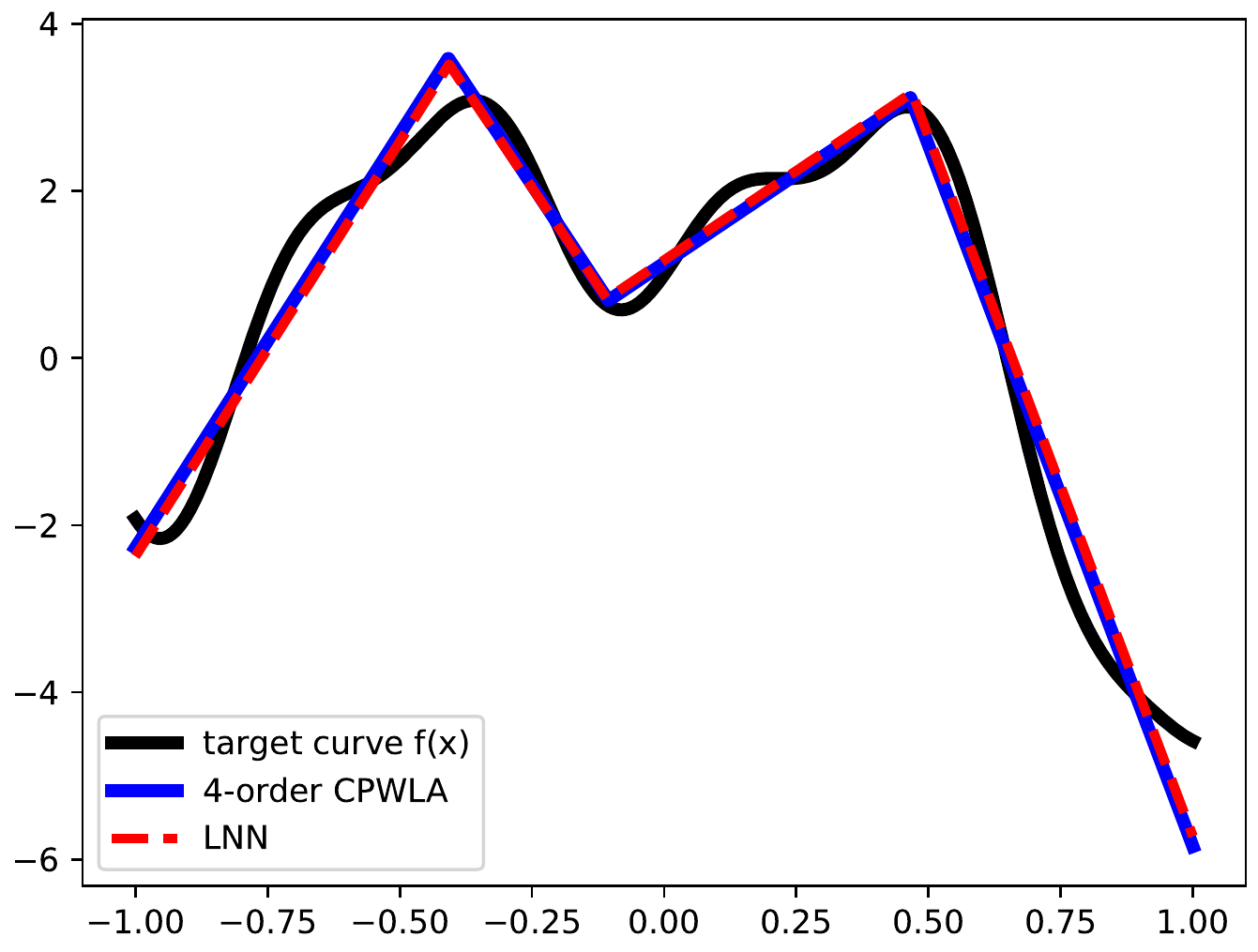}
\caption{$f(x)=5xsin(5x) + cos(5x)sin(10x)+exp(-x), x\in [-1,1]$}
\label{fig5c}
\end{subfigure}

\caption{Verify whether LNNs can find the global optimum.}
\label{fig5}
    \end{figure}

Can the back-propagation algorithm find the optimal solution even though the optimal solution of LNNs is the same as the corresponding CPWLA according to Eq.(\ref{eq16})? 

To answer the above question, we train three LNNs for the target functions listed in Table 1 respectively, whose results are displayed in Fig.\ref{fig5}. It is observed that LNNs do converge to the optimal solution, although there maybe exist a slight bias. While training LNNs, the most important thing is the selection of hyperparameters, which is crucial to the success of LNNs. 

As shown in Fig.\ref{fig5a}, we take the LNN with a single neuron and employ Adam optimizer with full-batch gradient descent by taking a learning rate as $3\times 10^ {-3}$. After training 300 epochs, the LNN was converged. The breakpoint of the trained LNN is at $x=0.0007$, which is quite close to the true breakpoint $x=0$.

In Fig.\ref{fig5b}, all the settings are the same as above, except that the epochs should be increased to 1500. It is noticed that the LNN finds another optimal solution, whose breakpoint is at $x=0.7072$. (The breakpoint of another true solution is at $x=-0.7071$).

Fig.\ref{fig5c} display the optimal solution found by the LNN with 3 hidden neuron, whose breakpoints are at $x=-0.4106$, -0.1054, and 0.4675. Compared with the breakpoints of the optimal solution $x=-0.4082$, -0.1055, and 0.4660, the LNN algorithm can indeed find a solution closer to the global optimum. To train the corresponding LNN, we apply SGD optimizer, take a learning rate as $5\times 10^ {-3}$, and optimize all the parameters until 2000 epochs with the batch size setting as 200.

\subsection{Comparison experiments}

\begin{figure}[pos=htpb]
    \centering
    \setkeys{Gin}{width=\linewidth,height=6cm} %set image parameters
\begin{subfigure}{8cm}
    \includegraphics{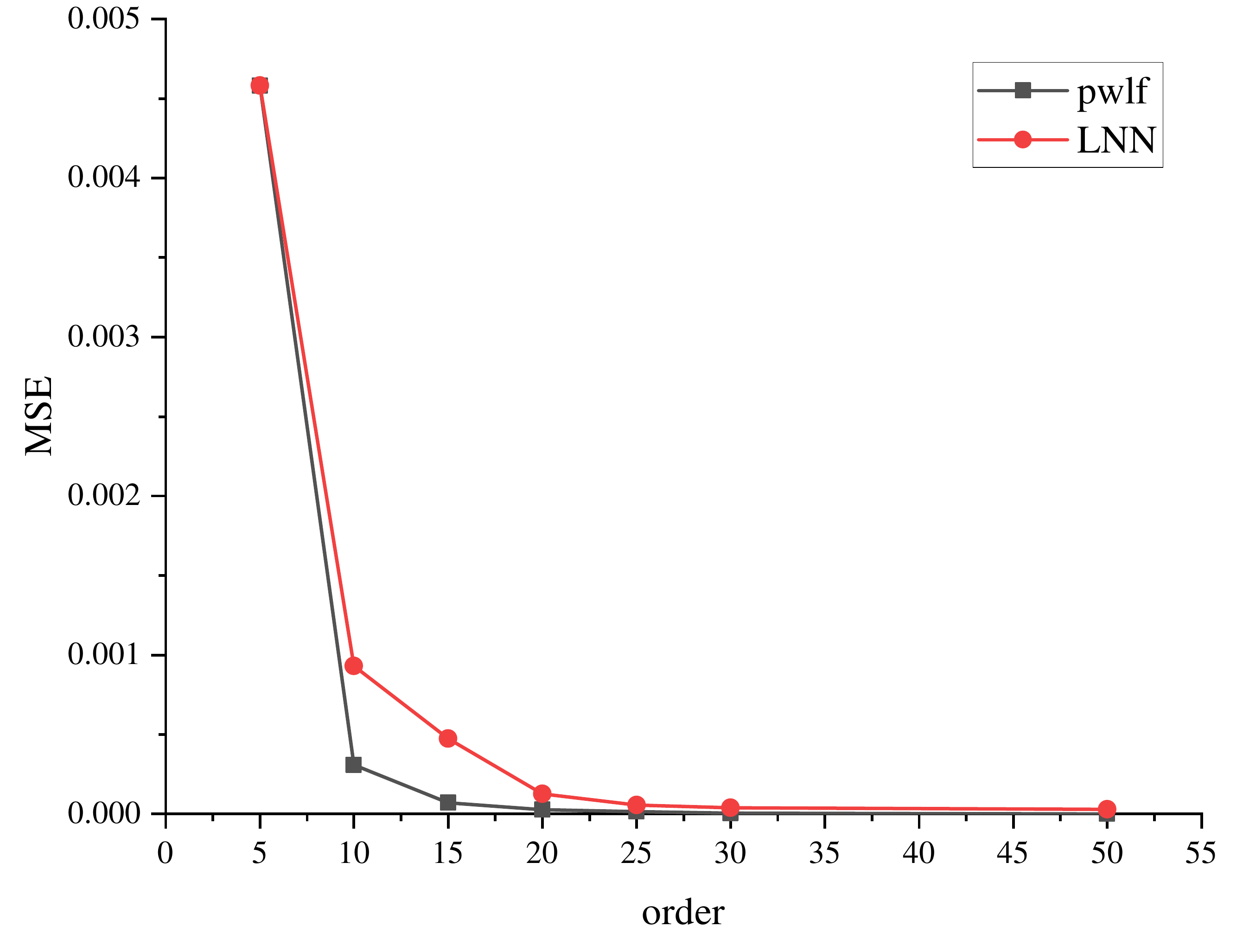}
    \caption{$f(x)=sin(\pi x)/{\pi x},x\in [-4,4]$}
    \label{fig6a}
\end{subfigure}
\hfil
\begin{subfigure}{8cm}
    \includegraphics{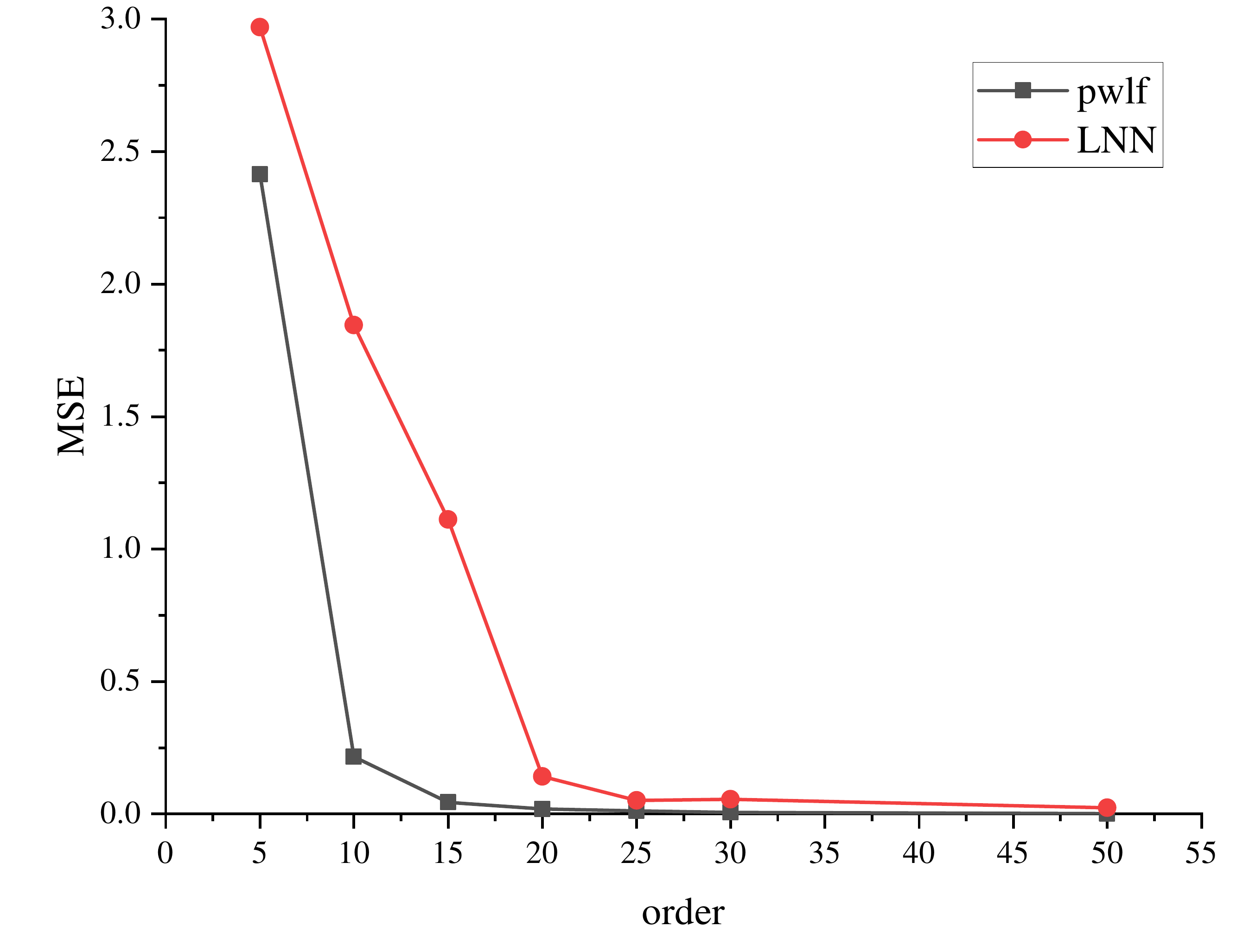}
    \caption{ $f(x)=sin(x)+xsin(x)cos(x),x \in [-10,10]$ }
    \label{fig6b}
\end{subfigure}

\medskip
\begin{subfigure}{8cm}
    \includegraphics{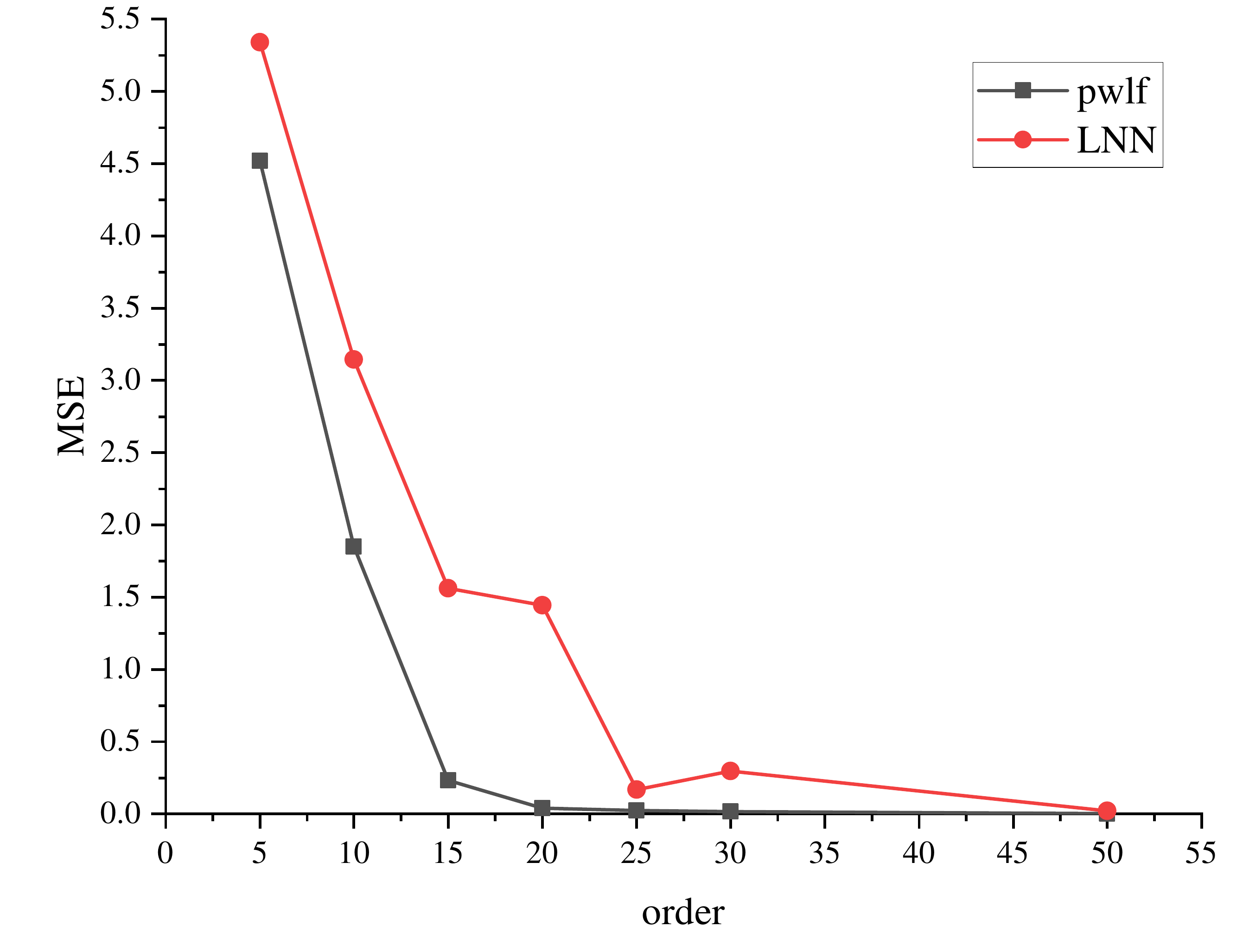}
\caption{$f(x)=20-5exp(-0.3x)-3exp(cos(\pi x)),x \in [-6,6]$}
\label{fig6c}
\end{subfigure}

\caption{The comparison of LNNs and the pwlf in terms of accuracy measured by MSE.}
\label{fig6}
    \end{figure}

\begin{figure}[pos=htpb]
    \centering
    \setkeys{Gin}{width=\linewidth,height=6cm} %set image parameters
\begin{subfigure}{8cm}
    \includegraphics{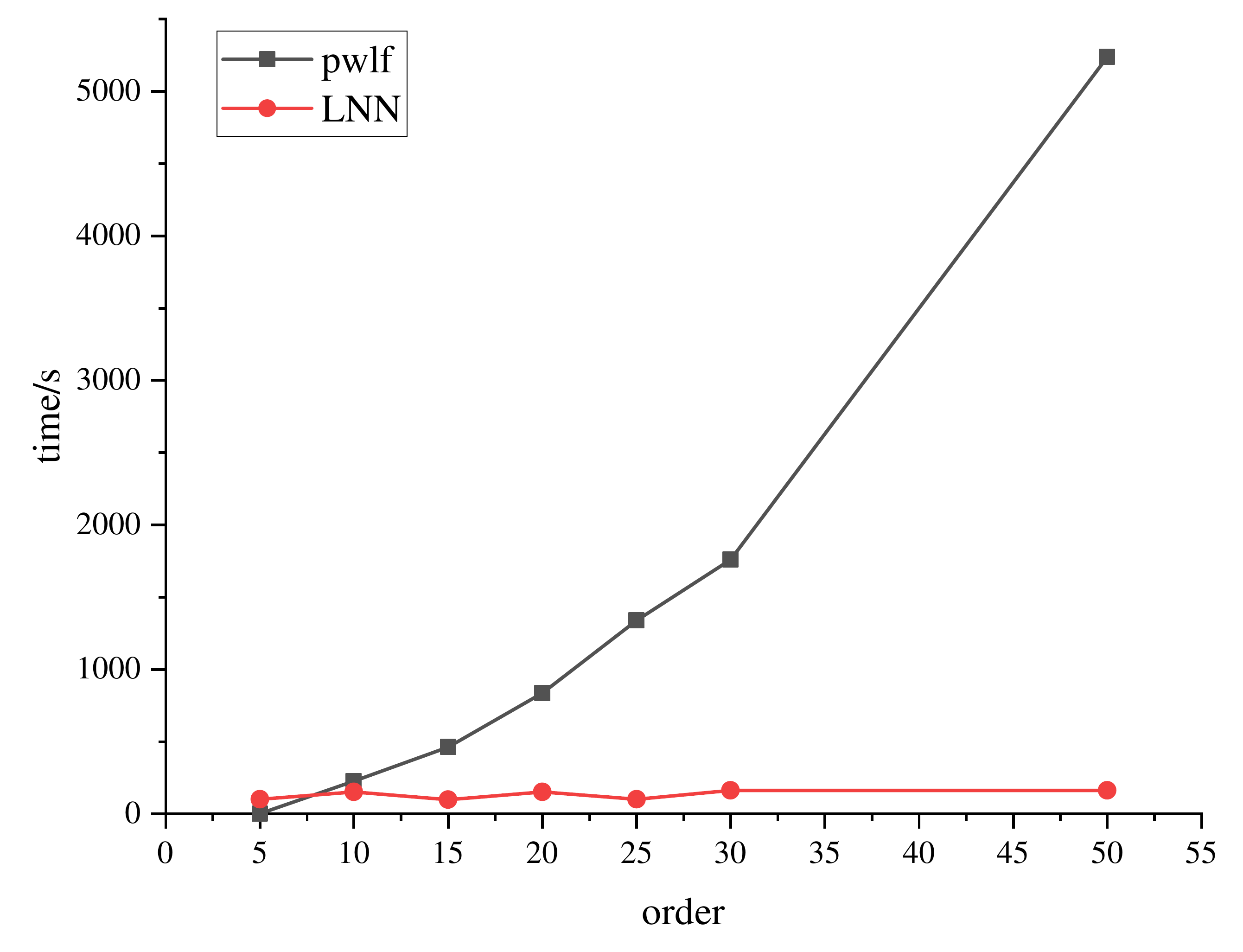}
    \caption{$f(x)=sin(\pi x)/{\pi x},x\in [-4,4]$}
    \label{fig7a}
\end{subfigure}
\hfil
\begin{subfigure}{8cm}
    \includegraphics{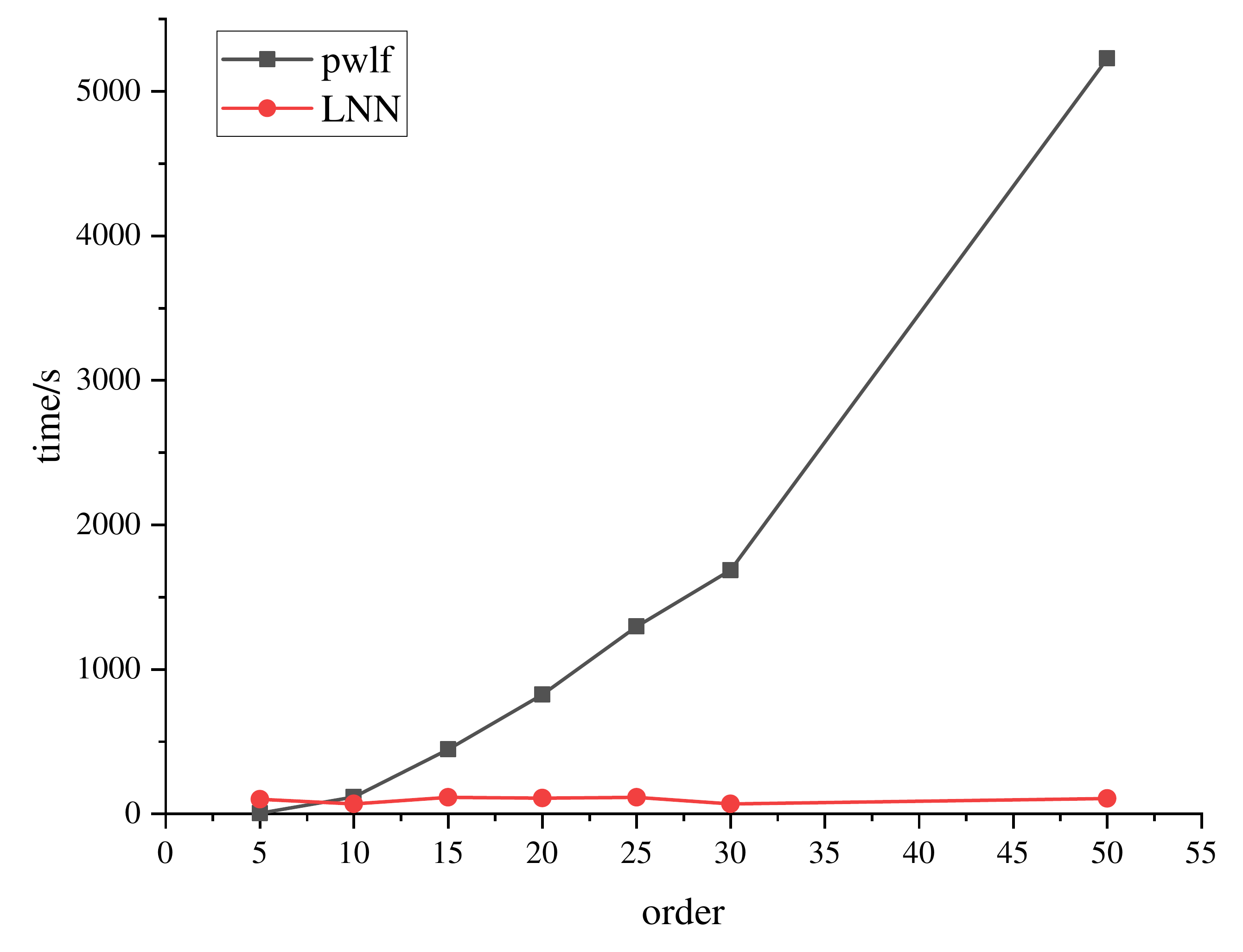}
    \caption{ $f(x)=sin(x)+xsin(x)cos(x),x \in [-10,10]$}
    \label{fig7b}
\end{subfigure}

\medskip
\begin{subfigure}{8cm}
    \includegraphics{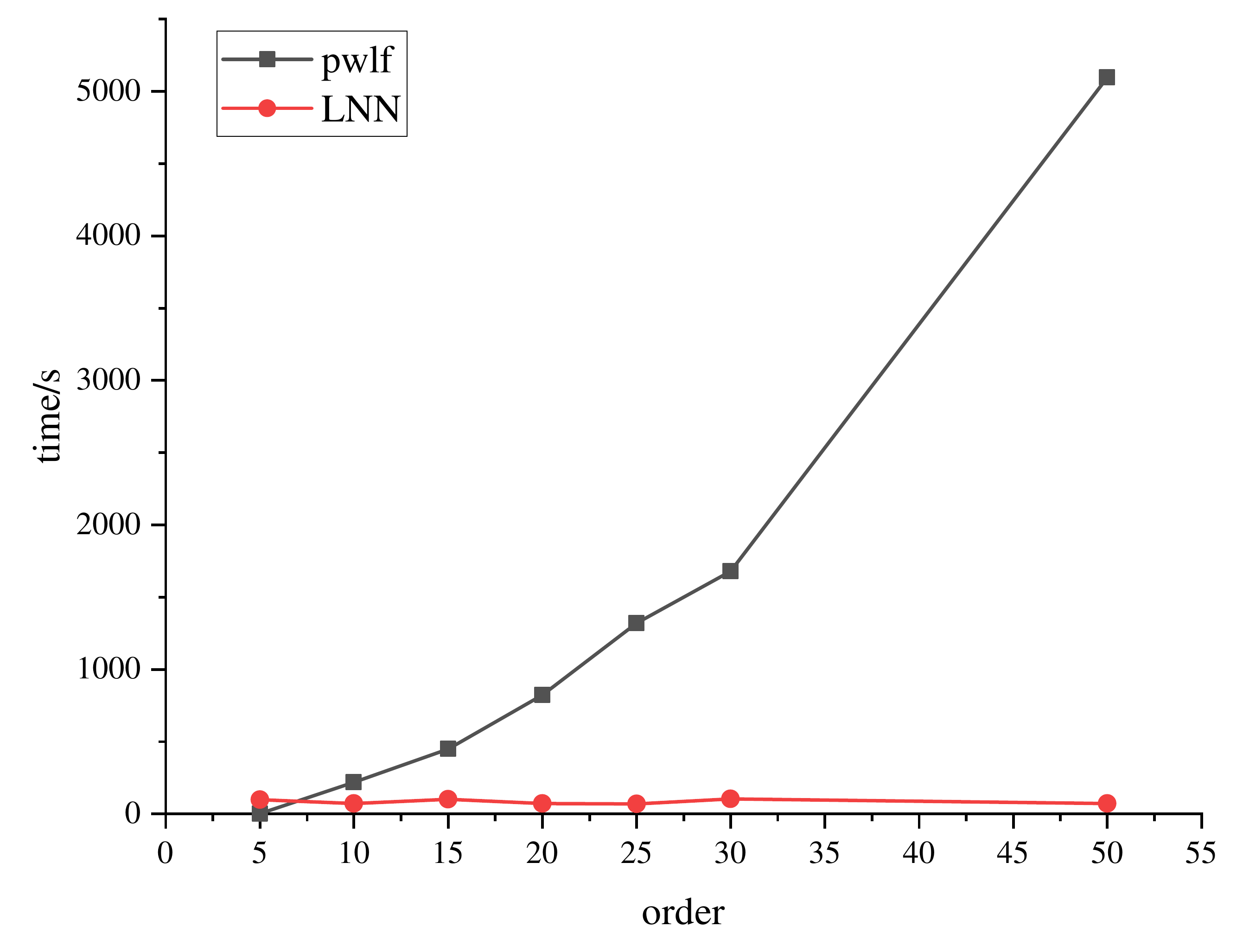}
\caption{$f(x)=20-5exp(-0.3x)-3exp(cos(\pi x)),x \in [-6,6]$}
\label{fig6c}
\end{subfigure}

\caption{The comparison of LNNs and the pwlf in terms of solving time.}
\label{fig7}
    \end{figure}

We evaluate the performance of LNNs and pwlf in terms of accuracy and time. To demonstrated intuitively, we depict the comparison between LNNs and pwlf algorithm, where Fig.\ref{fig6} shows the comparison with respect to the accuracy measured by the mean squares error (MSE), and Fig.\ref{fig7} shows the comparison about the solving efficiency measured by the time required. 

As shown in Fig.\ref{fig6}, the MSE decreases as the order increases. Compared to the pwlf, the MSE of LNNs is generally higher or comparable, which is consistent with the properties of neural networks. To ensure the LNNs well-trained, we determine the best learning rate and batch size from the candidate sets \{1e-3,5e-4,3e-4,1e-4,5e-5,3e-5,1e-5\} and \{20, 40, 100\} respectively.

As shown in Fig.\ref{fig7}, the time required for LNNs remains generally stable as the order grows, while the time required for the pwlf grows explosively. We should emphasize that the time required for LNNs refers to the time needed for a single training process instead of the time to search for all candidate parameters. Overall, our method is more adapted to the high-order CPWLA problem.

\subsection{Can LNNs approach the optimal solutions of CPWLA with fixed breakpoints?}

\begin{figure}
    \centering
    \setkeys{Gin}{width=8cm,height=6cm}
    \includegraphics{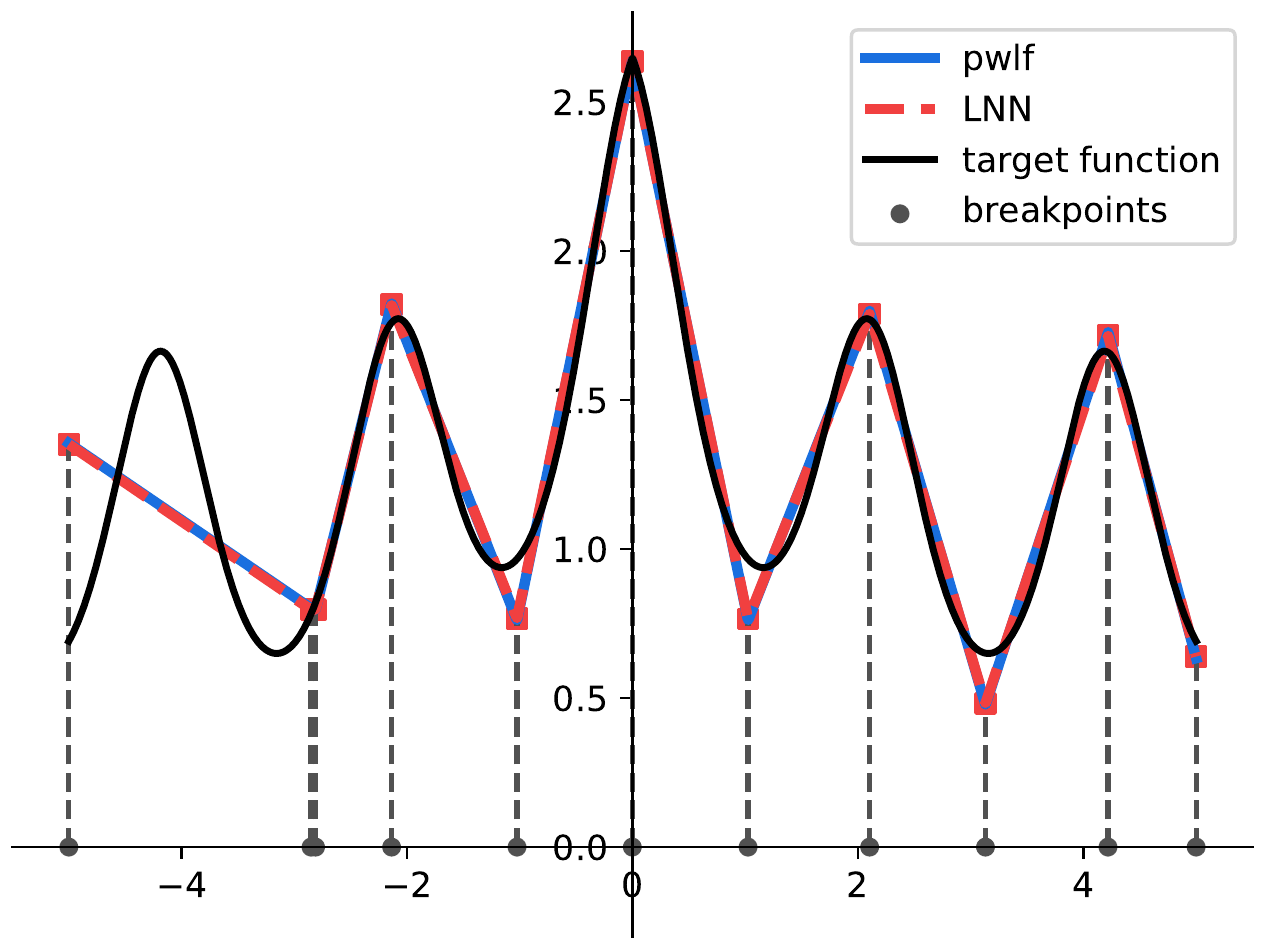}
    \caption{The comparison of the result of the LNN and the pwlf fitted with the same fixed breakpoints. The target function is  $f(x)=exp(-\sqrt{x^2}) + exp(0.5cos(3x)), x\in [-5,5]$. There are 2 points at $x_1=-2.8517$ and $x_2=-2.8114$, which are very close in the figure.}
    \label{fig8}
\end{figure}

The performance of LNNs depends on the configuration of hyperparameters. Furthermore, we wonder about the behavior of LNNs when the hyperparameters are not 'good'? It can be checked by comparing the well-trained LNNs with the least square solution in \cite{2019pwlf}. We design a target function $f(x)=exp(-\sqrt{x^2}) + exp(0.5cos(3x)), x\in [-5,5]$ and train a LNN with 9 hidden neurons. The learning rate is set as 3e-4, and batch size is taken by 20. In Fig.\ref{fig8}, we show the comparison of the result of the LNN and the pwlf fitted with the same fixed breakpoints (There are 10 straight lines in the figure, one of which is not clear because there are two points that are quite close). Consequently, we can conclude that the LNN can approach the optimal solutions of CPWLA with fixed breakpoints despite it is prone to fail in solving the global optimum.

\subsection{An empirical method to train LNNs}

It is palaver to search for suitable hyperparameters for LNNs. To deal with the difficulty, we found that the over-parameterization of LNNs generally leads to a laziness phenomenon in the performance, namely, as the order increases, the approximation accuracy grows more and more slowly. While the training of over-parameterized LNNs is more insensitive to the hyperparameters. Inspired by the phenomenon, we propose an empirical method to train LNNs. Firstly, for a given well-trained LNN, we train a new one with more hidden neurons to improve the performance, noted as $g(x)$. Then, we scan all the breakpoints $x_i$ except the endpoints of the interval and generate straight lines by connecting the point pairs $(x_{i-1},g(x_{i-1}))$ and $(x_{i+1},g(x_{i+1}))$. Finally, we calculate the MSE between the lines and the target function in turn, and eliminate the corresponding breakpoints with smaller MSE. Based on the filtered breakpoints and combined with the conclusion of Section 5.4, we can directly calculate the least squares solution.

\begin{figure}
    \centering
    \setkeys{Gin}{width=\linewidth,height=6cm} %set image parameters
\begin{subfigure}{8cm}
    \includegraphics{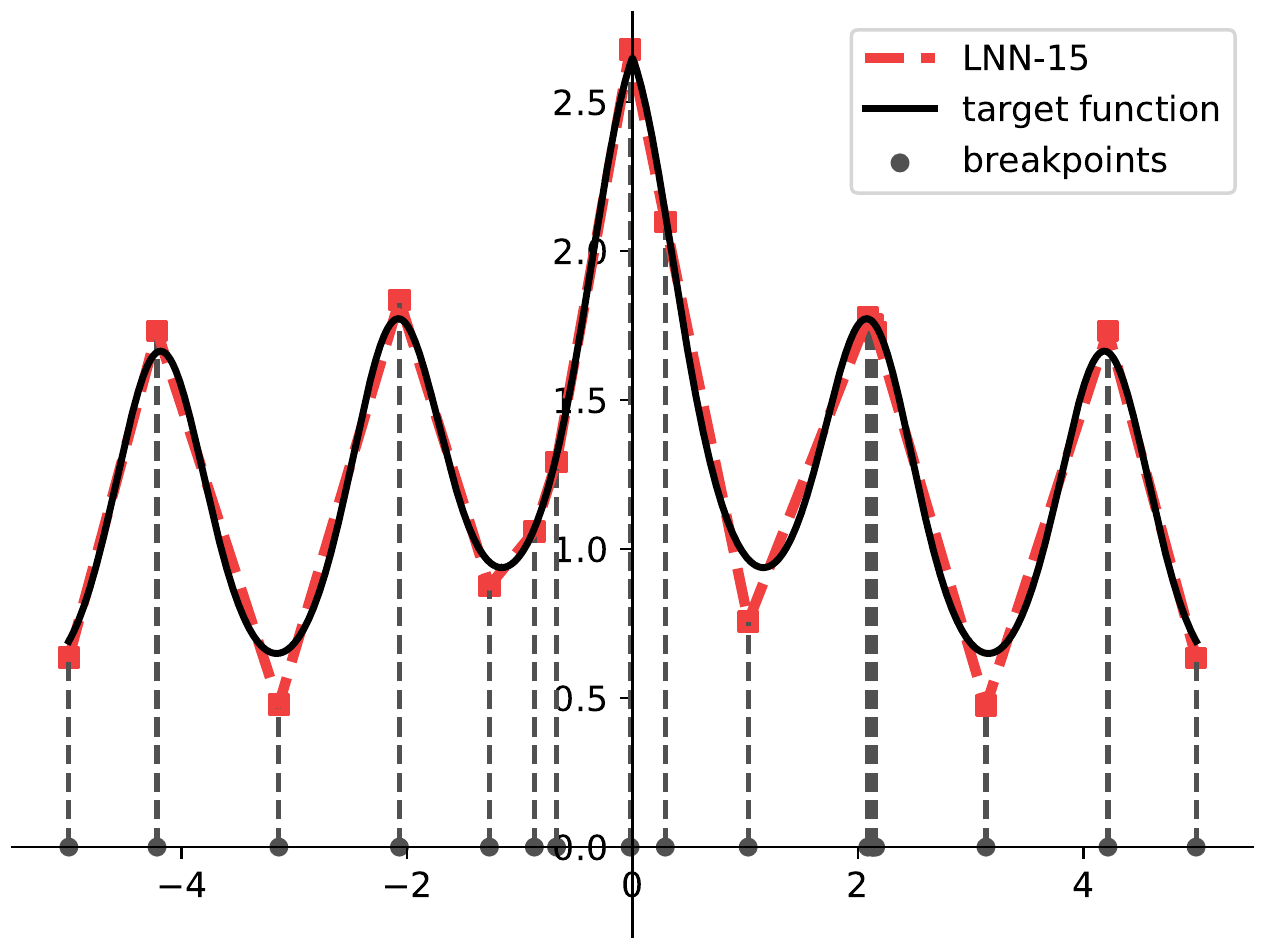}
    \caption{The result of the LNN with 15 hidden neurons.}
    \label{fig9a}
\end{subfigure}
\hfil
\begin{subfigure}{8cm}
    \includegraphics{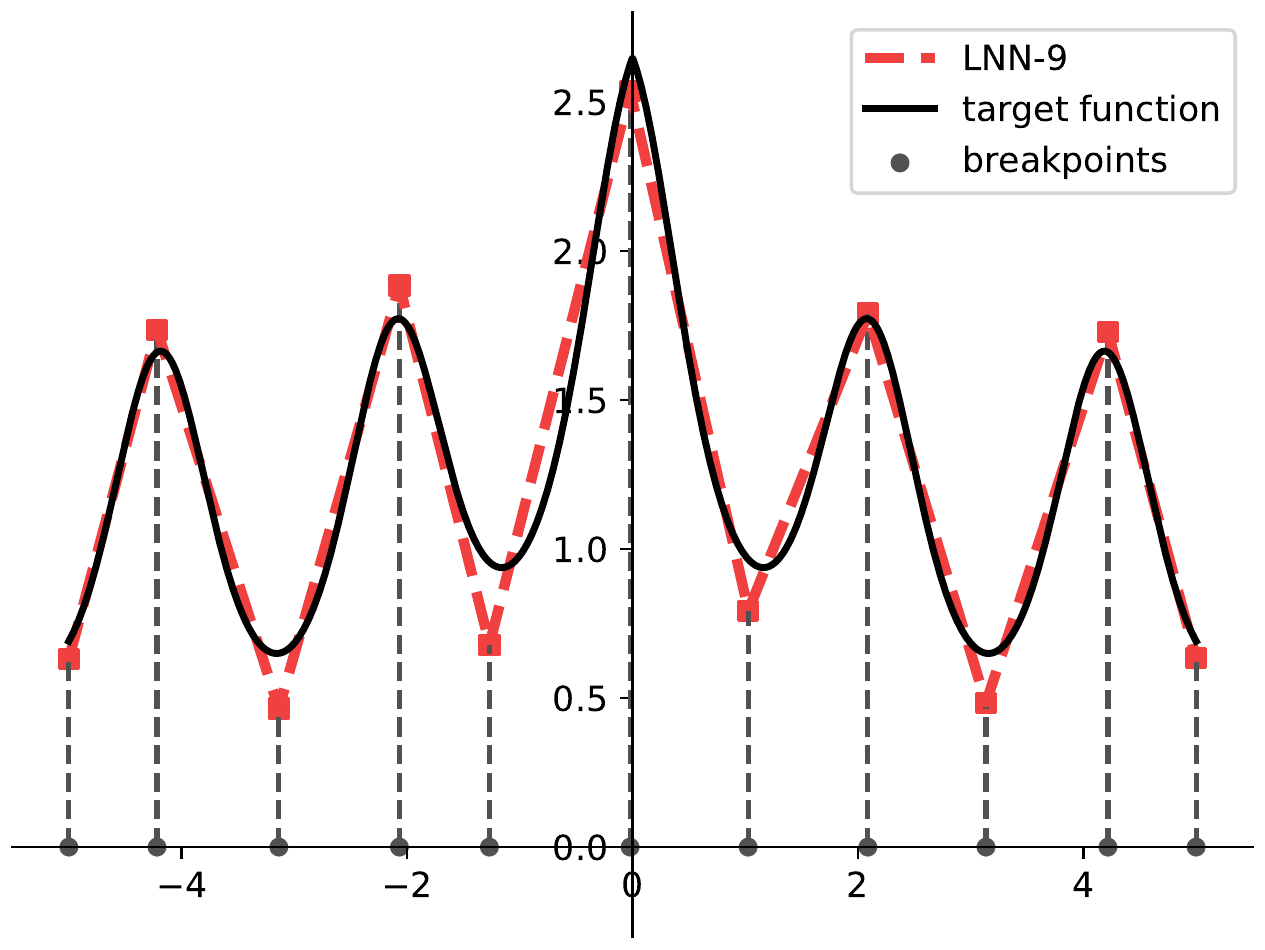}
    \caption{ The result of the pwlf fitted with 9 selected fixed breakpoints.}
    \label{fig9b}
\end{subfigure}

\caption{The evaluation of the proposed empirical method. The target funcrion is $f(x)=exp(-\sqrt{x^2}) + exp(0.5cos(3x)), x\in [-5,5]$}.
\label{fig9}
    \end{figure}

Following the above process, we retrain a LNN with 15 hidden neurons to solve the problem designed in Section 5.4, and select 9 breakpoints (except the endpoints $x_0=-5$ and $x_{10}=5$). Fig.\ref{fig9a}
shows the LNN with 15 hidden neurons, and Fig.\ref{fig9b} presents the final least squares result with selected fixed breakpoints. By comparing Fig.\ref{fig8} and Fig.\ref{fig9b}, we conclude that the proposed method does improve the performance of LNNs.

\section{Conclusion and future directions}

In this paper, we have studied the 1-D (C)PWLA problem and obtained the necessary and sufficient conditions of the optimal solution respectively. Moreover, we have associated the 1-D CPWLA problem with neural networks. We term the proposed neural network as the lattice neural network. In other words, the optimal solution of LNNs can be characterized by CPWLA, which can promote our understanding of deep learning. Meanwhile, the LNNs can be applied to solve the CPWLA problem. The experiments have shown that LNNs have significant advantages in solving time and comparable accuracy measured by the MSE. 

In the future, we will explore the LNN with high-dimensional input as well as more hidden layers. We are curious whether the proposed theorems still hold in high dimensions, which will provide a theoretical basis to understand the behavior of more complex neural networks.

\footnotesize

\nocite{*}
\bibliographystyle{elsarticle-num}
\bibliography{ref-work2.bib}

\vskip100pt

\makeatletter

\def\pct{\expandafter\@gobble\string\%}

\immediate\write\@auxout{\pct\space This is a test line.\pct }

\end{document}